\documentclass[dvipsnames,format=sigconf, language=english]{acmart}
\usepackage[ruled,vlined]{algorithm2e}
\usepackage{amsmath}
\usepackage{braket}
\usepackage{xcolor}

\usepackage{multirow}

\AtBeginDocument{%
  }

\copyrightyear{2024}
\acmYear{2024}
\setcopyright{rightsretained}
\acmConference[GECCO '24]{Genetic and Evolutionary Computation Conference}{July 14--18, 2024}{Melbourne, VIC, Australia}
\acmBooktitle{Genetic and Evolutionary Computation Conference (GECCO '24 Companion), July 14--18, 2024, Melbourne, VIC, Australia}
\acmDOI{10.1145/3638530.3664134}
\acmISBN{979-8-4007-0495-6/24/07}

\begin{document}

\title{Modeling stochastic eye tracking data: A comparison of quantum generative adversarial networks and Markov models}

\author{Shailendra Bhandari}
\email{shailendra.bhandari@oslomet.no}
\orcid{0000-0002-7860-4854}
\affiliation{%
  \institution{Department of Computer Science, OsloMet -- Oslo Metropolitan University, N-0130}
  \city{Oslo}
  \country{Norway}
}

\author{Pedro Lencastre}
\email{pedrog@oslomet.no}
\orcid{XXXX-XXXX-XXXX-XXXX}
\affiliation{%
  \institution{Department of Computer Science, OsloMet -- Oslo Metropolitan University, N-0130}
  \city{Oslo}
  \country{Norway}
}

\author{Pedro Lind}
\email{pedrolin@oslomet.no}
\orcid{0000-0002-8176-666X}
\affiliation{%
  \institution{Department of Computer Science, OsloMet -- Oslo Metropolitan University, N-0130 Oslo}
  \country{Norway;}
  \institution{Simula Research Laboratory, Numerical Analysis and Scientific Computing, 0164}
  \city{Oslo}
  \country{Norway}
}
\renewcommand{\shortauthors}{Bhandari et al.}

\begin{abstract}
 We explore the use of quantum generative adversarial networks QGANs for modeling eye movement velocity data. We assess whether the advanced computational capabilities of QGANs can enhance the modeling of complex stochastic distribution beyond the traditional mathematical models, particularly the Markov model. The findings indicate that while QGANs demonstrate potential in approximating complex distributions, the Markov model consistently outperforms in accurately replicating the real data distribution. This comparison underlines the challenges and avenues for refinement in time series data generation using quantum computing techniques. It emphasizes the need for further optimization of quantum models to better align with real-world data characteristics.

\end{abstract}

\maketitle

\section{Introduction}
The generative adversarial networks (GANs) \cite{goodfellow2014generative,10.1145/3422622} are the type of artificial intelligence algorithms that consist of two neural networks: the \emph{generator} and the \emph{discriminator}. The generator is trained to create realistic data, which are then used as negative instances for the discriminator to learn from. This process is also known as generative modeling, a branch of machine learning that involves training a model to produce new data similar to a given dataset. Meanwhile, the discriminator improves its ability to differentiate between authentic and generated data, providing feedback to the generator when it produces data that is not convincing. Throughout the training process, the generator continually strives to enhance its ability to create increasingly convincing forgeries. Simultaneously, the discriminator aims to improve its ability to discern between real and fake data accurately. Therefore, a GAN is a battle between two adversaries, the generator and the discriminator. The point of balance in this dynamic battle is achieved when the generator is capable of producing forgeries that are indistinguishable from the original training data, leaving the discriminator with a mere 50\% certainty in distinguishing real from fake outputs.

GANs become tremendously interesting and challenging topics in Machine Learning \cite{PavanKumar2020GenerativeAN}. It is beyond the boundary line of computational creativity, showcasing increasingly amazing examples each year \cite{liao2020unsupervised}. While GANs have shown remarkable power and interest, it is also limited by various challenges: difficulties in achieving stable training of the GAN \cite{10386654, orponen1994computational}, the vanishing gradient, and the mode collapse \cite{arjovsky2017wasserstein}. Efforts from both the academic community and the commercial sector have been made to address these problems \cite{garg2020advances, wiebe2015quantum,PhysRevResearch.2.033212,doi:10.1126/sciadv.aav2761,stein2022quclassi}. The exploration of Quantum Deep Learning models has been gaining popularity recently, followed by the concept of quantum supremacy, which Google has showcased \cite{Arute2019}, along with the potential for a quantum edge in machine learning. Recently, discussions around Quantum Generative Adversarial Networks (QGANs) mainly focused on applications like efficient data handling or seeking benefits over traditional classical models \cite{zoufal2019quantum}. An example of a successful QGAN involves using a manageable number of quantum bits to train the GAN to recognize and replicate random patterns. The motivation behind quantum-enabled research is the potential quantum advantage that could partially alleviate the computational complexity issue and search for the best solution to the mode collapse and vanishing gradient problem. 

In the context of integrating quantum principles into neural networks, it is important to recognize foundational work that has set the stage for contemporary advancements. One of the important works is by Behera et al.  \cite{Behera2005}, who developed a recurrent quantum neural network (RQNN) model to describe eye movements in tracking moving targets. Their work utilized a nonlinear Schrödinger wave equation to mediate the neural responses, providing a foundation for understanding how quantum mechanics can be integrated with neural processing. This early exploration into quantum neural networks is important. Our approach to QGANs aims to enhance the generative capabilities of neural networks through the incorporation of quantum bits, quantum gates, and the fundamental properties of quantum mechanics: superposition and entanglement.

The main goal of this paper is to use the QGAN and test its performance in reproducing stochastic trajectories and compare them with some benchmarks from stochastic modeling. Previously, Lencastre et al. \cite{LENCASTRE2023133831} tested the classical GANs'performance, comparing with a mathematical model, namely a Markov chain. They found that GANs struggle to capture rare events and cross-feature relations and are unable to create faithful synthetic data successfully. Here, we test the QGANs performance compared with the previous classical GANs and mathematical models. One of the problems of the classical GAN is the vanishing gradient problem when generating discrete data \cite{LENCASTRE2023133831} to overcome this issue we explore the ability of quantum computing, which naturally has advantages to equip GANs with the ability to deal with discrete data.
\section{The Quantum GAN model}
\begin{figure}
    \centering
    \includegraphics[width=0.75\linewidth]{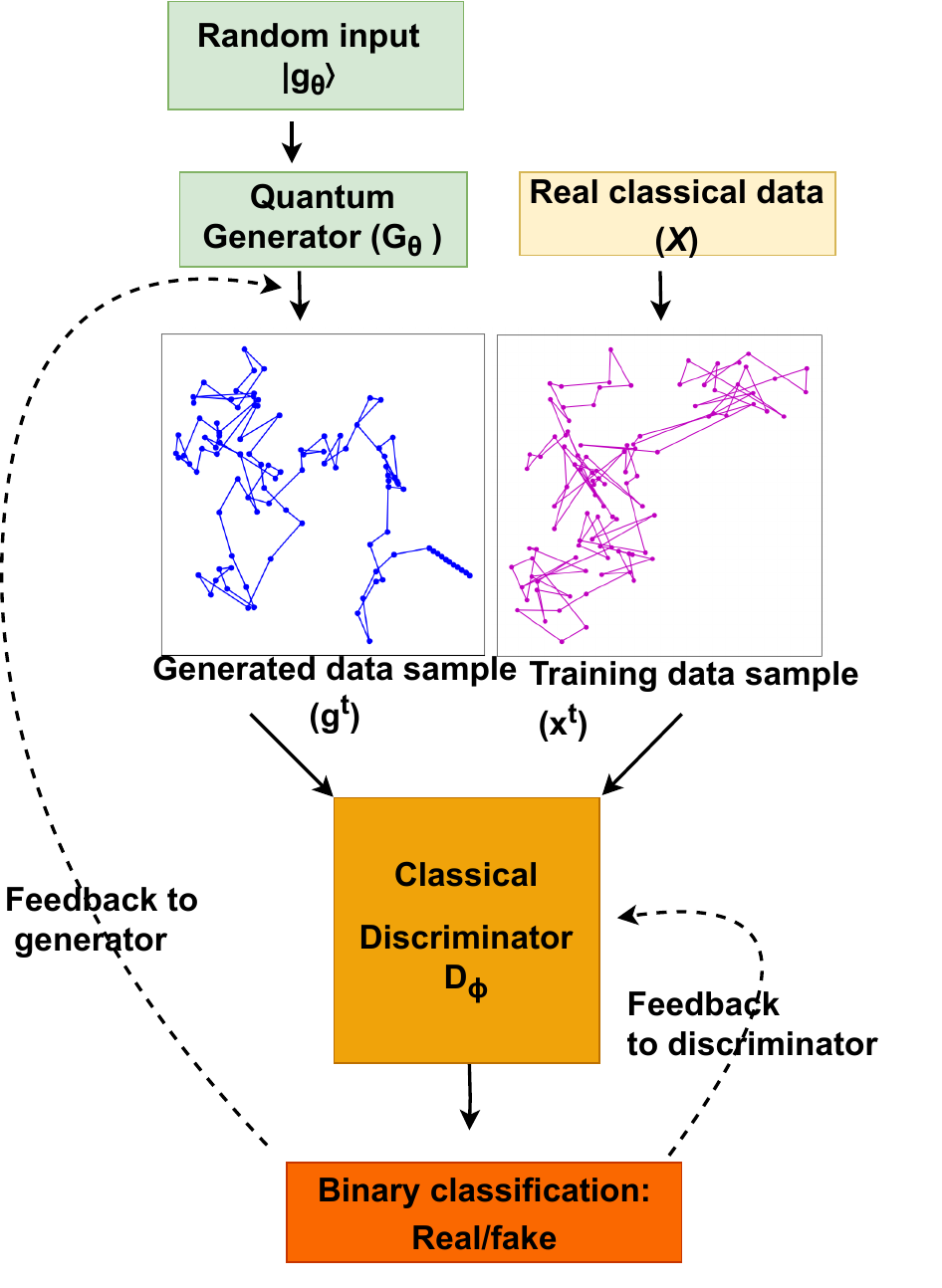}
    \caption{Generative adversarial network models workflow: the generator generates data samples ($g_t$) to imitate the real-world data and tries to fool the discriminator. The discriminator differentiates the generated and the training data samples by training both the generator and discriminator alternatively until the loss converges towards the Nash equilibrium.}
    \Description{GAN workflow.}
    \label{fig:GAN_architecture}
\end{figure}

In our study, the utilized GAN framework comprises two neural networks: the generator and the discriminator, as illustrated in Figure \ref{fig:GAN_architecture}. These networks undergo alternating training phases. Consider a classical training data set \(X = \{x^0, ..., x^{s-1}\}\) drawn from unknown time series distribution. The generator (\(G_\theta\)) receives a random noise vector \(z\) and then produces the generated sample \(G(z)\). The discriminator \(D_ \phi\) is trained to distinguish between the training data \(x\) and the generated data \(G(z)\). The parameters of \(D_ \phi\) are updated in order to maximize \cite{SITU2020193}:

\begin{equation}
\mathbb{E}_{x \sim P_{d}(x)}[\log D_\phi(x)] + \mathbb{E}_{z \sim P_{z}(z)}[\log(1 - D_\phi(G_\theta(z)))]\; ,
\end{equation}
where \(P_d(x)\) is the real-time series distribution and the \(P_z(z)\) is the distribution of the input noise. The output \(D_\phi(x)\) can be explained as the probability that \(D_\phi\) thinks the sample \(x\) is real. The goal of the discriminator here is to maximize the probability of correctly classifying to make \(D_\phi\)  a better adversary so that \(G_\theta \) has to try harder to fool the \(D_\phi\). Similarly, the parameters of the \(G_\theta\) are updated to maximize: 
\begin{equation}
\mathbb{E}_{z \sim P_{z}(z)}[\log (D_\phi(G_\theta(z)))]\; , 
\end{equation}
to convince \(D_\phi\) that the generates samples \(G(z)\) are real. This process is illustrated in Figure~\ref{fig:GAN_architecture}. 

The goal of optimizing classical GANs can be approached from several perspectives. In this study, we adopt the non-saturating loss function \cite{fedus2018paths}, which is also implemented in the original GAN publication's code \cite{NIPS2014_5ca3e9b1}. The generator loss function is given by:
\begin{equation}
L_G(\theta) = -\mathbb{E}_{z \sim P_{z}(z)}[\log (D_\phi(G_\theta(z)))]\; ,
\end{equation}
which aims at maximizing the likelihood that the generator creates samples that are labeled as real data samples. Also, the discriminator's loss function is given by the expression:
\begin{equation}
L_D(\phi) = \mathbb{E}_{x \sim P_{d}(x)}[\log D_\phi(x)] + \mathbb{E}_{z \sim P_{z}(z)}[\log(1 - D_\phi(G_\theta(z)))]\; ,
\end{equation}
which aims at maximizing the likelihood that the discriminator labels training data samples as training data samples and generated data samples as generated data samples. In practice, the expected values are approximated by batches of size \( m \)
\begin{equation}\label{OptimizerGan}
L_G(\phi, \theta) = -\frac{1}{m} \sum_{i=1}^{m} [\log D_{\phi}(G_{\theta}(z^{(i)}))],
\end{equation}
and
\begin{equation}\label{optimizerDisc}
L_D(\phi, \theta) = \frac{1}{m} \sum_{i=1}^{m} [\log D_{\phi}(x^{(i)}) + \log (1 - D_{\phi}(G_{\theta}(z^{(i)})))],
\end{equation}
for \( x' \) in \( X \) and \( z' \) sampled from \(P_d(x)\). 


\subsection{Quantum generator}

\begin{figure*}
    \centering
    \includegraphics[width=1\linewidth]{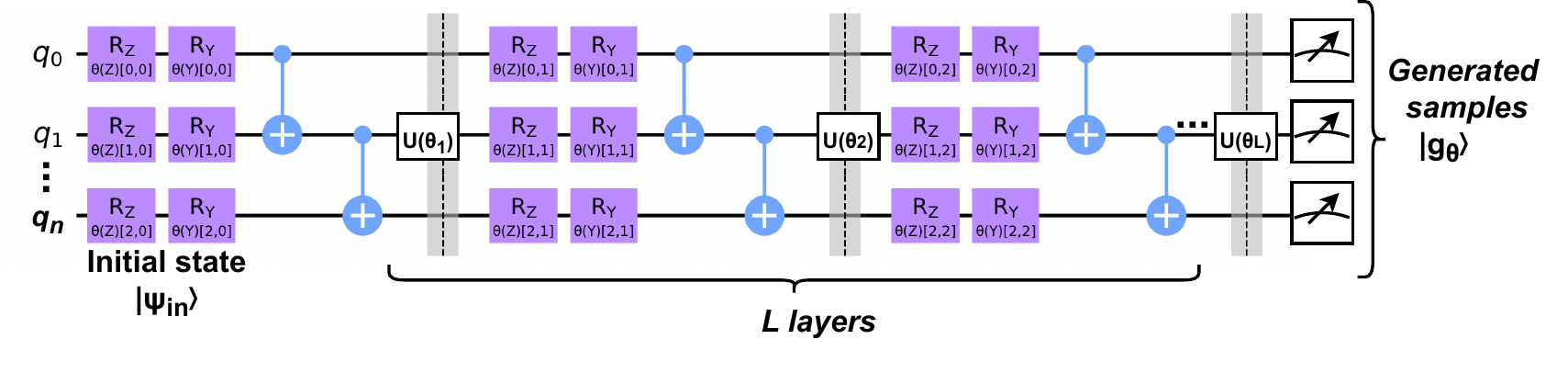}
    \caption{The quantum generator circuit in variational form with \(L\) layers acting on \(n\) qubits. Each layer in the circuit is composed of single qubit rotation gates (\(R_Z, R_Y\)) and two-qubit controlled-phase gates.}
    \Description{GAN}
    \label{fig:qganCircuit}
\end{figure*}

The model we designed is a hybrid architecture (quantum-classical) where the input is in classical form with a classical discriminator and generates an output by measurement of a state from the parametrized quantum circuit. Specifically, it is a quantum GAN that generates a classical discrete time series, which is composed of a parametrized quantum circuit as a generator and a classical feed-forward neural network as a discriminator. 

The parameterized quantum circuit is then expressed as \( G(\boldsymbol{\theta}) \), where \( \boldsymbol{\theta} = \{\theta_1, \ldots, \theta_k\} \) represents the set of parameters which can be tuned as necessary. This circuit is operated by the single qubit \( R_Y \) and \( R_Z \) rotation gates, along with the two-qubit Controlled-NOT (CX) gates, all operating on an initial state \cite{10.1007/978-3-031-43085-5_53}. When applying the rotation operators on each qubit, it is expressed as:
\begin{equation}
    \prod_{i=1}^{N} R_Z^i(\theta^i _{l,2}) R_Y^i(\theta^i _{l,1}),
\end{equation}
where \(l\) denotes the \(l^{th}\) layer and \(i\) denotes the \(i^{th}\) qubit. \( R_z \) and \( R_Y \) are the rotation gates given by following expressions: 
\begin{equation}
\begin{split} 
R_Z(\theta) = e^{-i\theta\sigma_z/2} = \cos{\frac{\theta}{2}I}- i\sin{\frac{\theta}{2}\sigma_z} = \begin{bmatrix}
e^{-i\theta/2}&0\\
0&e^{i\theta/2}
\end{bmatrix}\; ,\\
R_Y(\theta) = e^{-i\theta\sigma_y/2} = \cos{\frac{\theta}{2}I}- i\sin{\frac{\theta}{2}\sigma_y} = \begin{bmatrix}
\cos{\frac{\theta}{2}}&-i\sin{\frac{\theta}{2}}\\
i\sin{\frac{\theta}{2}}&\cos{\frac{\theta}{2}}
\end{bmatrix}.
\end{split}
\end{equation}

The Controlled-NOT gate is used to create entanglement between the qubits, and this process is expressed as:
\begin{equation}
    \prod_{i=1}^{N} CU^i_{(i\;mod\;N)+1}\; ,
\end{equation}
where \(i\) denote the controlled qubit and \((i\;mod\;N)+1\) represent the target qubit. The number of single-qubit gates represents the number of parameters in the generative quantum circuit in a circuit which is \(3N\) per layer.

The parametrized quantum circuit for generating \(N-bit\) samples consists of \(N\) qubit and \(U(\theta_L)\) layers as shown in Figure \ref{fig:qganCircuit}. The input quantum state \(\ket{\psi_{in}}\) is initialized to \(\ket{0}^{\otimes N}\) and passed through \(L\) layers of unitary operator. All the qubits outcomes of the circuit are measured on the computational basis at the end of the circuit. The measurement outcome is then collected to form \(N\)-bit sample \(X\). Overall, the quantum generator is trained to transform \(N\)-qubit input to \(N\)-qubit output state \cite{zoufal2019quantum}:
\begin{eqnarray}
    \ket{g_\theta} & =& G_\theta \ket{\psi_{in}}\cr
    & =& \prod_{p=1}^{2^N} \left( \bigotimes_{q=1}^{N} (R_Y R_Z (\theta^{q,p})) U(\theta_L) \right) \bigotimes_{q=1}^{N} (R_Y R_Z (\theta^{q,0}) \ket{\psi_{in}})\cr
    & = &\sum_{j=0}^{2^N - 1} \sqrt{p_j} \ket{j} \; , \label{Qgan}
\end{eqnarray}
where samples are drawn by measuring the output state \(\ket{g_\theta}\) in computational basis with measurement outcomes \(\ket{j}\), \(j \in \{0,1....,2^N-1\}\). \(N\) is the number of qubits. In addition, the term \(p_j\) is the sampling probability associated with the basis state $\ket{x_j}$. For \(m\) data samples (\(g^l\)) from the quantum generator and \(m\) randomly chosen training data samples (\(x^l\)), where \(l = 1,...m\), the generator and discriminator are optimized with a respective loss function.

The generator's goal is to generate data that can fool the discriminator. The generator loss function for a data batch of size \(m\) is expressed as:
    \begin{equation}
    L_G(\phi, \theta) = -\frac{1}{m} \sum_{i=1}^{m} [\log D_{\phi}(g^l)],
    \end{equation}
or equivalently,
\begin{equation}
    L_G(\phi, \theta) = \sum_{j=0}^{2^N-1} p_\theta ^j \log (D_{\phi}(g^j)),
\end{equation}
where \(p_\theta ^j = |\bra{j}{g_\theta}\rangle|^2\).
The loss function of the discriminator is given by
    \begin{equation}
        L_D(\phi,\theta) = \frac{1}{m}\sum_{l=1}^m [log D_\phi(x^l)+log(1-D_\phi(g^l))].
    \end{equation}

Gradient-based optimization techniques can enhance convergence speed, particularly near local optima in a convex region, when compared to methods that do not utilize gradient information \cite{PhysRevLett.126.140502}. This method for calculating analytical gradients \cite{farhi2018classification, zoufal2019quantum, PhysRevA.99.052306, PhysRevA.98.062324} for the variational circuit is as discussed below. The parameters $\boldsymbol{\theta}$ can be updated with gradient-based methods that require the evaluation of
\begin{equation}\label{gradient_based_method}
\frac{\partial \mathcal{L}_G (\phi, \boldsymbol{\theta})}{\partial \theta^{i,l}} = - \sum_{j=1}^{m} \frac{\partial p^j_\theta}{\partial \theta_{i,l}} \log (D_\phi (g^{j})) . 
\end{equation}

Eq. \eqref{gradient_based_method} can be further evaluated based on Ref. \cite{farhi2018classification}
\begin{equation}\label{gradient_based_analytical}
\frac{\partial p_j^\theta}{\partial \theta^{i,l}} = \frac{1}{2} \left( p_{\theta^{i,j}_+}^j - p_{\theta^{i,j}_-}^j \right) , 
\end{equation}
with $\theta^{i,l}_\pm = \theta^{i,l} \pm \frac{\pi}{2} e_{i,l}$ and $e_{i,l}$ denoting the $(i, l)$-unit vector of the respective parameter space.


The selection of parameters \( \boldsymbol{\theta} \) is important, especially when \( l > 1 \). A key factor to consider is the circuit depth, as increasing it enhances the complexity of the quantum circuit. Utilizing a circuit depth greater than \( 1 \) is advantageous for training on complex datasets, as deeper circuits can capture more intricate structural patterns. Thus, it is essential to design a circuit that is both deep and rich in parameters. This choice is driven by the need to effectively capture and represent the complex distribution characteristics of the data. By increasing both the number of parameters and the circuit's depth, we can significantly enhance the generator's ability to accurately model the data. Such a strategy is pivotal for quantum generators tasked with processing and generating data distributions of considerable complexity.

\subsection{The discriminator}
The discriminator model, a classical feed-forward neural network, is constructed and implemented using PyTorch's \cite{paszke2019pytorch} neural network module. It comprises a three-layer LSTM \cite{10.1162/neco.1997.9.8.1735} (Long Short-Term Memory) network with 128 hidden units and bidirectional processing. This LSTM feeds into a series of linear layers that further process the data. A dropout of 0.3 is applied after the first linear transformation to prevent overfitting. The output of the discriminator is a single scalar value representing the probability that the input data is from the real distribution, obtained through a sigmoid activation function.

The discriminator is trained using AMSGRAD \cite{reddi2019convergence} with a learning rate of 0.002 and a momentum of 0.999. Using the first and the second momentum terms, AMASGRAD is a robust optimization technique for non-stationary objective functions and noisy gradients \cite{kingma2017adam}. The loss function employed is binary cross-entropy (BCE) and is suitable for binary classification tasks such as distinguishing between real and fake data sequences. During training, the discriminator evaluates both real and generated data sequences, updating its weights to minimize the loss function. The training stability is maintained using a gradient penalty on a discriminator's loss function \cite{10386654, Kodali2018OnCA}. The analytic computation of the quantum generator loss function gradients is implemented based on Equations. \eqref{gradient_based_method} and \eqref{gradient_based_analytical}. The training procedure is iteratively conducted for a specified number of epochs, with both generator and discriminator losses monitored to ensure the model's convergence.

The adversarial training of the quantum GAN described above is illustrated in Algorithm 1. The training of the GAN iterates for the specified number of epochs, until the convergence of the loss function. At each epoch, the loss function optimizes alternatively to parameters \(\theta\) and \(\phi\).

\begin{algorithm}[t] 
    \label{Algorith1}
    \SetAlgoLined
    \textbf{1. Input:} Eye-tracking velocity 
    distribution, hyperparameters\;
    \textbf{2. Initialization:}\\
    2.1. Load and preprocess eye-tracking data\;
    2.2. Set seed for reproducibility across quantum and classical computations\;
    2.3. Initialize quantum circuit for the generator with specified qubits and an EfficientSU2 ansatz\;
    2.4. Initialize a PyTorch-based classical neural network that represents the classical discriminator\;
    2.5. Define adversarial loss function for training\;
    2.6. Define optimizer for both generator and discriminator according to Equations \eqref{OptimizerGan} and \eqref{optimizerDisc} with specified learning rates and hyperparameters\;
    \textbf{3. Training} QGAN (generator and discriminator)\;
    \For{each epoch}{
        \For{each batch in training data}{
            a. Generate fake data using the quantum generator\;
            b. Compute discriminator loss on both real and generated data, backpropagate and update discriminator\;
            c. Generate fresh fake data and compute generator loss against the discriminator's feedback, backpropagate, and update generator\;
            d. Record losses for analysis and model tuning\;
        }
        e. Save generator and discriminator losses for the current epoch\;
    }
    \textbf{4. Output:} Generator and discriminator losses;
    visualize training progression with generator and discriminator losses\;
    \caption{QGAN training for eye-tracking data}
\end{algorithm}


\section{Markov-chain models of time-series}

We generate a two-dimensional time series based on a Markov transition matrix $T$, where each element $T_{ij}$ denotes the transition probability from state $i$ to state $j$. Mathematically, a time series \(X_t\) is said to follow a Markov process if it fulfills a Markov property:
\begin{equation}
\begin{split}
Pr(X_{t=j} = \hat{x}_j | X_{t=j-1} = \hat{x}_{j-1}, \ldots, & X_{t=0} = \hat{x}_0) = \\
& Pr(X_{t=j} = \hat{x}_j | X_{t=j-1} = \hat{x}_{j-1}),
\end{split}
\end{equation}
for all positive integers j, where capital letters mean stochastic
variables at different time steps and lowercase letters are the respective values of those variables.

In a Markov model, time-series generation is based on computing the conditional probability $Pr(X_{t=n+1} = x_{n+1} \,|\, X_{t=n} = x_{n})$. This is estimated empirically with the Gaussian estimation kernel $K$ as follows:
\begin{equation}
    Pr(X_{t=n+1} = x_{n+1} \,|\, X_{t=n} = x_{n}) = \frac{Pr(X_{t=n+1} = x_{n+1}, X_{t=n} = x_{n})}{Pr(X_{t= n} = x_{n})},
\end{equation}
with
\begin{equation}
\begin{split}
    Pr(X_{t = n+1} = x_{n+1}, X_{t=n} = x_{n}) = & \frac{1}{(\hat{N} - 1)h^2} \sum_{i=1}^{\hat{N}-1} K\left(\frac{x_{n+1} - \hat{x}_{i+1}}{h}\right) \\
    & \times K\left(\frac{x_{n} - \hat{x}_{i}}{h}\right),
\end{split}
\end{equation}
and the kernel $K$ defined as
\begin{equation}
    K\left(\frac{x_{n} - \hat{x}_{i}}{h}\right) = \frac{1}{\sqrt{2\pi}} \exp \left(-\frac{1}{2} \left(\frac{x_{n} - \hat{x}_{i}}{h}\right)^2 \right).
\end{equation}
The bandwidth $h$ of the Gaussian estimation kernel is computed following Silverman's rule \cite{silverman1998density}:
\begin{equation}
    h = \left(\frac{4\hat{\sigma}^5}{3\hat{N}}\right)^{\frac{1}{5}} \approx 1.06 \hat{\sigma} (\hat{N} - 1)^{-1/5},
\end{equation}
where $\hat{\sigma}$ is the standard deviation of the sample and $N$ is the number of data points in our sample.

For empirical data analysis, the conditional probability can be represented by a transition matrix $T$ of dimension $N_s \times N_s$ with entries
\begin{equation}
    T_{ij} = Pr(X_{t=n+1} \in [k_i, k_{i+1}) | X_n \in [k_j, k_{j+1})),
\end{equation}
with $i, j \in N, i, j \in [0, N_s)$ and $k_m > k_n$ for $m > n$.

The efficacy of the proposed method is contingent upon three critical factors: (i) the robustness of the Markov assumption \cite{RUSSO2013822}; (ii) the number of states \(N_s\), which directly influences computational efficiency and scales with \(N_s^4\) for bi-dimensional processes; and (iii) the sample size \(N\), which affects the bandwidth \(h\) and thereby determines the spatial resolution of the model. We implemented this process using the open source code accessible via GitHub \cite{LENCASTRE1}.


\section{Data and the statistical measure}
We used the Eye-tracking data\footnote{All data collected was anonymized and follows the ethical requirements from the Norwegian Agency for Shared Services in Education and Research (SIKT), under the application with Ref. 129768.} which was gathered at Oslo Metropolitan University utilizing the advanced eye-link Duo device, capable of reaching up to 2000 Hz but was adjusted to 200 Hz for this study. The measurements, recorded in screen pixels, were taken as participants searched for specific targets in images from the book ``Where's Waldo?'' \cite{bookwaldo}. Each of the eight selected images was viewed for two minutes, a duration not expected to suffice for finding all targets but intended to keep participants focused. The data from eye-tracking measurements were preprocessed and utilized to train a QGAN. Initially, the velocity data for both left and right eyes are calculated by finding the Euclidean distance between consecutive position points and then dividing by the time interval, set at 1/200 seconds, to convert this distance into velocity. This velocity data is then aggregated into a structured format suitable for feeding into the discriminator. The data undergoes resampling to a fixed interval of 10 seconds, aggregating measurements by their mean within these intervals to reduce noise and temporal variability. The resultant dataset is then normalized using a MinMaxScaler to ensure the data fits within the required operational range of  0 and 1 \cite{DEAMORIM2023109924}. The normalized data is transformed into sequences of a specified length (100 data points in this case), which are then used to create training batches. These sequences were fed into the discriminator, where it assessed them and learned to discern features that distinguish real data from the outputs generated by the quantum generator. This process aided the discriminator in enhancing its accuracy in identifying authentic samples.

The performance of the QGAN model is measured by the Jensen-Shannon (JS) divergence \cite{e21050485}. This metric, a symmetrized variant of the Kullback-Leibler (KL) divergence \cite{kullback1951information}, offers a symmetric distance measure between probability distributions. The JS divergence \cite{weng2019gan} is defined by 
\begin{equation}
D_{JS}(P | Q) = \frac{1}{2} D_{KL}(P | M) + \frac{1}{2} D_{KL}(Q | M),
\end{equation}
where $P$ and $Q$ are distributions, and \( M = \frac{1}{2}(P + Q) \). The KL divergence is standard for assessing distributional similarity which enhances maximum likelihood estimates. The JS divergence, while preserving these properties, is more intuitive as it assesses the approximation of synthetic distributions to empirical ones. Therefore JS divergence in this case can be relevant for discriminators within QGANs to distinguish synthetic data from the generator.


\section{Results}

\begin{figure*}
    \centering
    \includegraphics[width=1\linewidth]{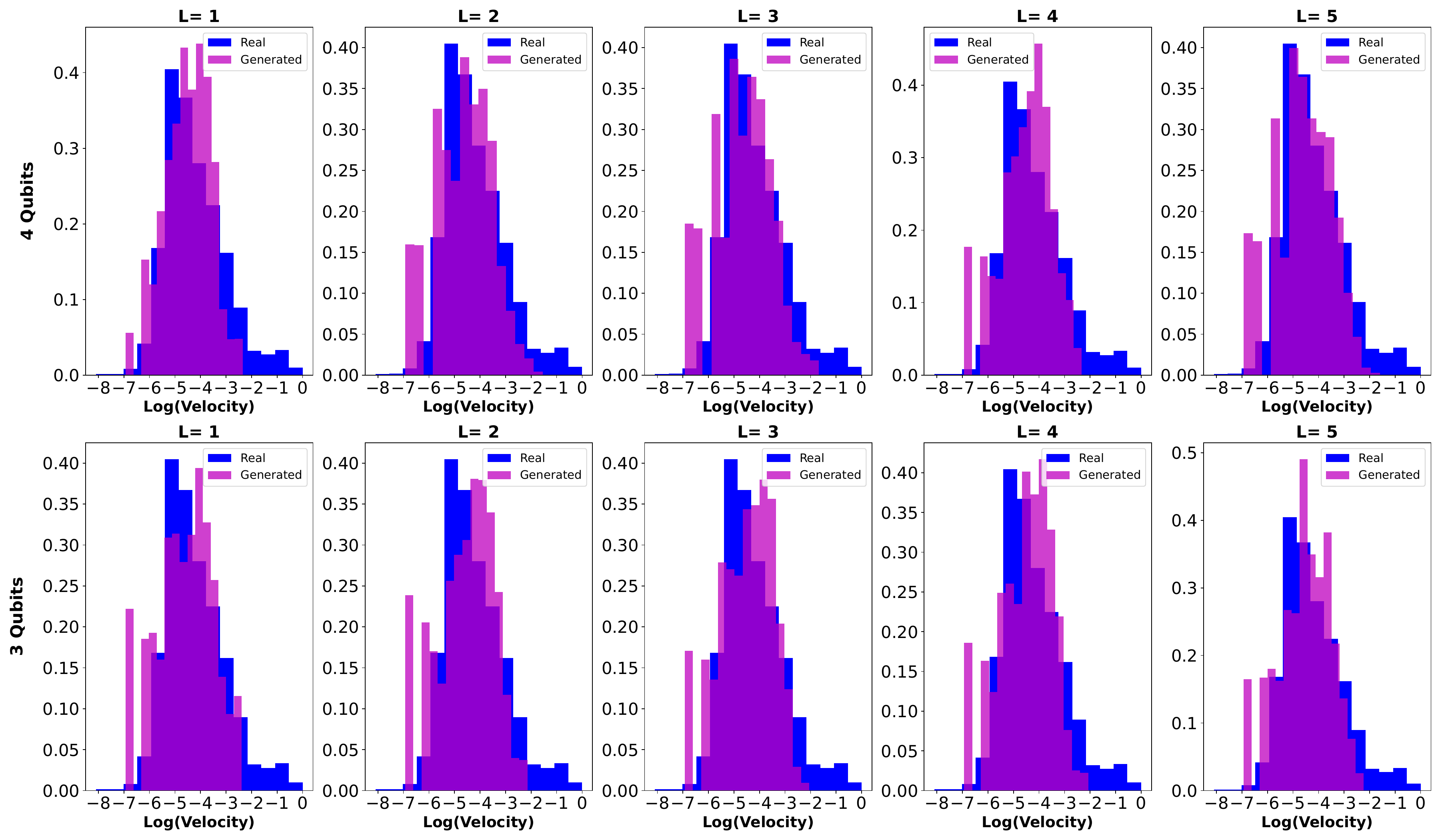}
    \caption{Comparative histograms of log-transformed real and generated eye movement velocity data across different circuit layers (1 - 5) for three and four-qubit QGAN. Each subplot illustrates the distribution of real data (in blue) with the corresponding generated data (in magenta) at respective circuit layers. The use of log transformation ensures a focus on the distribution's dynamics rather than its absolute scale to facilitate a clearer understanding of the model's performance across varying complexities.}
    \Description{ Real data is in red and generated data is in blue.}
    \label{fig:realVsGenerated_data3Qubit}
\end{figure*}

The implementation of QGAN was done using PyTorch \cite{paszke2019pytorch} and IBM's Qiskit \cite{Qiskit} simulator 0.45. The discriminator's LSTM network is built with linear layers and a dropout mechanism distinguished between real and generated data sequences. The quantum generator is modeled by a parametrized quantum circuit and was optimized to replicate the complex distribution of the eye movement velocities.  For the QGAN applied to eye-tracking data, the discretization and representation in quantum states was done using 3 and 4 qubits, each offering a different resolution of data representation. With 3 qubits, the velocity data is mapped onto $2^3 = 8$ discrete levels, resulting in the quantum state:
\begin{equation}
G_\theta \ket{\psi_{in}} = \sum_{i=0}^{7} \sqrt{p_{\theta}^i} |i\rangle,
\end{equation}
where $|i\rangle$ represents the $i$-th discretized state of the velocity data. $p_{\theta}^i$ are the associated probabilities, learned through the training of the quantum generator as elaborated in Eq.~\eqref{Qgan}. The training of our QGAN model was carried out over 300 epochs, with data batches of 500 samples each. 

Figures \ref{fig:realVsGenerated_data3Qubit} illustrate the histograms of log-transformed real versus generated data for three and four-qubit configurations across five layers. The use of log transformation ensures a focus on the distribution's dynamics rather than its absolute scale to facilitate a clearer understanding of the model's performance across varying complexities.

The evaluation of configurations of 3 and 4 qubits presents a nuanced understanding of the models' capabilities in reproducing real eye movement velocity data. Tab.~\ref{tab:statistics_combined} shows the statistical analysis for the generated data across different circuit layers including the Markovs model and the real data. The data generated by the QGANs exhibits a trend of increased skewness and kurtosis with increasing layers in quantum circuits, indicating a divergence from the real data's distribution especially in distribution tails. The generated distributions to the actual data are further quantified through the JSD values, as shown in Tab.~\ref{tab:jsd_results}. For 3 qubits QGAN performance is relatively stable across layers. For 4 qubits the second and third layers show optimal JSD, indicating minimal divergence points. Notably the Markov model exhibits superior performance with the lowest JSD, highlighting its efficiency in achieving minimal divergence compared to the QGAN-generated data.

\begin{figure}
    \centering
    \includegraphics[width=1\linewidth]{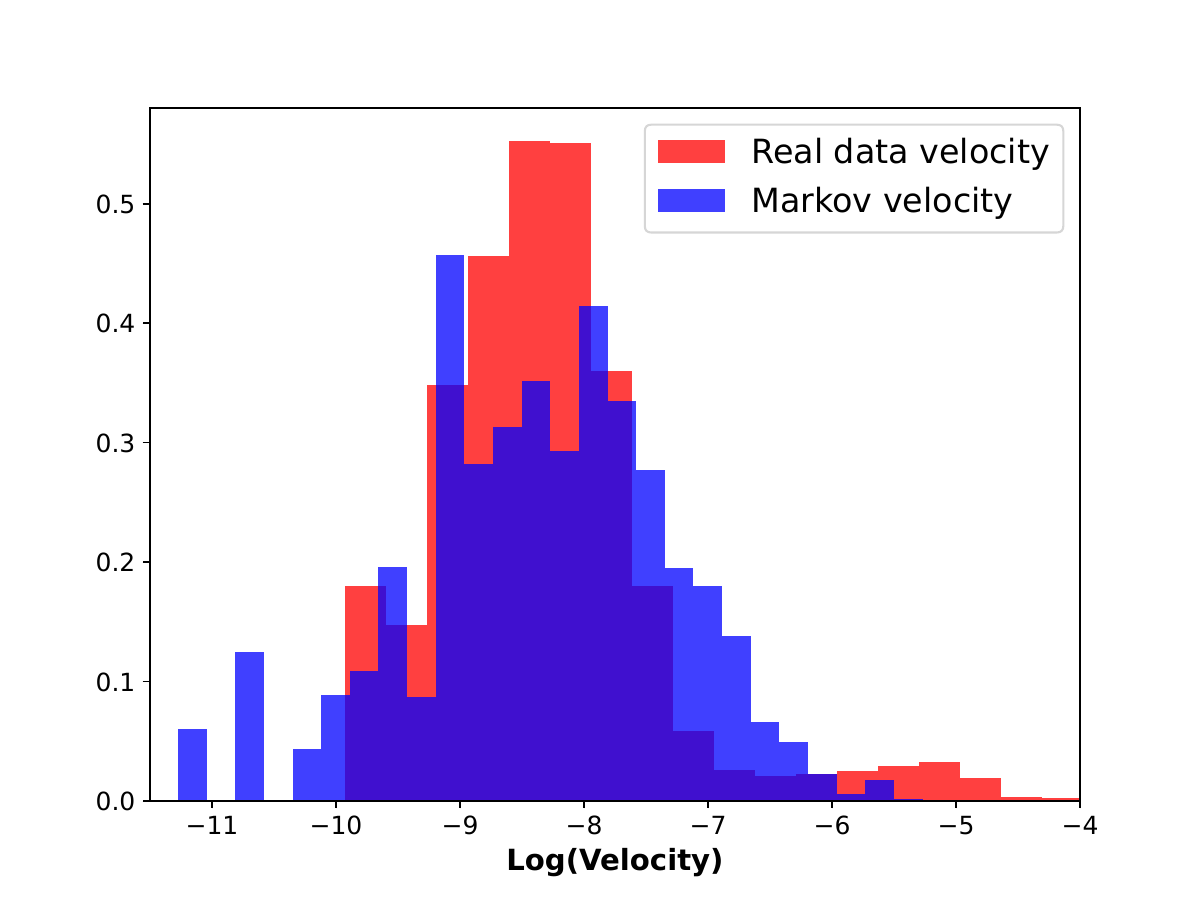}
    \caption{Comparative histograms of log-transformed eye movement velocities: real versus Markov model.}
    \Description{Comparative histograms of log-transformed eye movement velocities: real versus Markov model.}
    \label{fig:realVsGenerated_Markov}
\end{figure}

The comparison to a mathematical Markov model, with its significantly lower JSD value, highlights the distinct challenges faced in quantum data generation. Figure \ref{fig:realVsGenerated_Markov} illustrates the comparative histograms of log-transformed eye movement velocities between real data and the generated data by a Markov model. The visual comparison and the lower JSD value clearly emphasize the effectiveness of Markov models in simulating complex data distributions closely resembling real data.


\begin{table}
\centering
\caption{Comparison of generated data for 3 and 4 qubits with real data}
\label{tab:statistics_combined}
\begin{tabular}{@{}lcccc@{}}
\toprule
Layers [3 Qubits] & Mean & Std Dev & Skewness  & Kurtosis \\ \midrule
1 - Generated & 7.1196e-05 & 0.00131 & 26.0450 & 933.951 \\
2 - Generated & 7.1196e-05 & 0.00147 & 30.7205 & 1253.51 \\
3 - Generated & 7.1196e-05 & 0.00120 & 41.7435 & 2740.014 \\
4 - Generated & 7.1196e-05 & 0.00180 & 54.2916 & 4773.553 \\
5 - Generated & 7.1196e-05 & 0.00183  & 60.87   & 6174.75 \\

\midrule
Layers [4 Qubits] & Mean  & Std Dev & Skewness  & Kurtosis \\ \midrule

1 - Generated & 7.1196e-05 & 0.0012 & 22.964 & 674.157 \\
2 - Generated & 7.1196e-05 & 0.0013 & 28.720 & 1117.93 \\
3 - Generated & 7.1196e-05 & 0.0014 & 32.326 & 1743.243 \\
4 - Generated & 7.1196e-05 & 0.0015 & 37.681 & 2377.453 \\
5 - Generated & 7.1196e-05 & 0.0015 & 38.528 & 2641.427 \\

\midrule
\textbf{Real data} & 0.001735 & 0.04855 & 9.19 & 128.26 \\
\midrule
\textbf{Markov model} & 0.000371 & 0.00044752 & 3.493265 & 19.6143 \\
\bottomrule
\end{tabular}
\end{table}


\begin{table}
\centering
\caption{Jensen-Shannon divergence for 3 and 4 qubits across depths compared with a Markov model.}
\label{tab:jsd_results}
\begin{tabular}{ccc}
\toprule
Layers (L) & 3 Qubits & 4 Qubits \\
\midrule
1 & 0.08940 & 0.080354 \\
2 & 0.08895 & {0.080198} \\
3 & 0.08885 & {0.080153} \\
4 & 0.08872 & {0.080196}\\
5 & 0.08859 & {0.080102} \\
\midrule
Markov model &  & 0.00034021547 \\
\bottomrule
\end{tabular}
\end{table}

\section{Discussion and conclusions}

The comprehensive analysis of QGANs alongside Markov models, particularly in modeling stochastic eye movement velocity, underscores the promising aspects of quantum computing in handling complex datasets. However, when we put side by side with the mathematical Markov model, our analysis paints a comprehensive picture of the present state of synthetic data generation via quantum methods. 

We have shown that by increasing the number of qubits as well as the number of layers, QGANs improve their performance in reproducing eye-gaze trajectories.
However, 
despite QGANs' innovative nature and their capability to discern intricate eye gaze data patterns, Markov models outshine them in closely mirroring the actual data distributions.

This observation not only reinforces the previously identified challenges faced by AI in creating synthetic data—a topic already explored within classical GANs \cite{LENCASTRE2023133831} research, but it also broadens this understanding to include the emerging domain of quantum computing. Such findings underscore the critical need for ongoing advancements in quantum algorithm development. The goal is clear: to overcome these obstacles and fully harness quantum computing's potential to produce synthetic datasets that closely resemble their real-world analogs.

\section{Acknowledgments}
We extend our sincere gratitude to Mr. Ramesh Uprety for his invaluable contributions and insightful discussions on the architecture of GANs, hyperparameter tuning strategies, and careful data-cleaning processes. We are deeply appreciative of his collaborative spirit and expert advice, which have significantly enriched this work. This work was funded by the Research Council of Norway under grant number 335940 for the project `Virtual-Eye'.

\bibliographystyle{ACM-Reference-Format}
\bibliography{reference}

\end{document}